\documentclass[letterpaper, 10pt, conference]{IEEEtran}  % Comment this line out
\IEEEoverridecommandlockouts                              % This command is only
\usepackage[utf8]{inputenc}
\usepackage[T1]{fontenc}
\usepackage{amsmath, amssymb}
\usepackage{graphicx, float}
\usepackage{enumitem}
\usepackage{tabularx,booktabs}

\newcolumntype{C}{>{\centering\arraybackslash}X} % centered version of "X" type
\title{Interpreting the Latent Space of Generative Adversarial Networks using Supervised Learning}

\author{
\IEEEauthorblockN{Toan Pham Van$^{1}$, Tam Minh Nguyen$^{1}$, Ngoc N. Tran$^{1}$, Hoai Viet Nguyen$^{1}$,\\ Linh Bao Doan$^{1}$, Huy Quang Dao$^{1}$, Thanh Ta Minh$^{1,2}$}
\IEEEauthorblockA{$^{1} $R\&D Lab, Sun* Inc\\
\{pham.van.toan, nguyen.minh.tamb, tran.ngo.quang.ngoc, nguyen.viet.hoai,\\ doan.bao.linh, dao.quang.huyb, ta.minh.thanh\}@sun-asterisk.com\\
$^{2} $Le Quy Don Technical University, 236 Hoang Quoc Viet, Bac Tu Liem, Ha Noi\\
thanhtm@mta.edu.vn}
}

\begin{document}

\maketitle
\thispagestyle{empty}
\pagestyle{empty}

%%%%%%%%%%%%%%%%%%%%%%%%%%%%%%%%%%%%%%%%%%%%%%%%%%%%%%%%%%%%%%%%%%%%%%%%%%%%%%%%
\begin{abstract}

With great progress in the development of Generative Adversarial Networks (GANs), in recent years,  the quest for insights in understanding and manipulating the latent space of GAN has gained more and more attention due to its wide range of applications. While most of the researches on this task have focused on unsupervised learning method, which induces difficulties in training and limitation in results, our work approaches another direction, encoding human’s prior knowledge to discover more about the hidden space of GAN. With this supervised manner, we produce promising results, demonstrated by accurate manipulation of generated images. Even though our model is more suitable for task-specific problems, we hope that its ease in implementation, preciseness, robustness, and the allowance of richer set of properties (compared to other approaches) for image manipulation can enhance the result of many current applications.

\end{abstract}


%%%%%%%%%%%%%%%%%%%%%%%%%%%%%%%%%%%%%%%%%%%%%%%%%%%%%%%%%%%%%%%%%%%%%%%%%%%%%%%%
\section{INTRODUCTION}

The task of image generation has introduced many interesting applications in the world of computer vision, including image-to-image translation \cite{brock2018large, isola2016imagetoimage}, character drawing generation \cite{jin2017automatic}, and more. In recent years, a lot of effort has been done into enhancing Generative Adversarial Networks (GAN) \cite{goodfellow2014generative} - a model that produced very promising results for the above applications. The main objective of this model is to generate realistic data from a random vector, in much lower dimension, sampled from a prior distribution. Called the ``feature vector'', this random vector is believed to have encoded and condensed the important characteristics of the image; and the task of GAN is to generate an image from that information.
% wait ??? The latent space does contain any information. Its the generator that learn the mapping from a point in that space to another point in data distribution

As the quality of image samples increases, more attention has also been drawn to latent space interpretation and image manipulation. This task studies how interactions between uni-variables in the latent space $z$  of GAN results in generated images that we observe, particularly how to sample images with desired properties or manipulate attributes of synthesized images. The task can be usefully applied in Photograph Editing \cite{perarnau2016invertible}, Face Aging \cite{8296650},…

Overall, many works on learning the latent space share a main limitation: they do not encode human’s prior knowledge into their studies. If one has some prior knowledge about the domain, they can appropriately assume several factors of variation of the data. Taking human faces as an example, it's common to presume that a successful GAN model’s latent space can encode numerous attributes including: faces pose, hair’s colors, attractiveness, baldness, smiling... This knowledge can be useful for task-specific training. Many approaches can be applied more generally; however, they can be either very time-consuming and difficult to train (such as InfoGAN \cite{chen2016infogan}) or unable to learn specific factors of variation at all (for instance, vector arithmetic \cite{radford2015unsupervised} and interpolation \cite{radford2015unsupervised} between two images can only show that latent space indeed encodes meaningful semantics).

In this paper, we propose a method of learning the hidden latent space, and as a direct result, meaningfully manipulating GAN model‘s generated images. Specifically, since the human prior knowledge is in the form of labels provided with the data, we shall call the domain of these features the label space of the input data, which we explore to learn a linear mapping between the latent space and its own attributes (i.e., the labels). The main idea is to control image generation by carefully manipulating the latent vectors through label variables.

Theoretically, we expect a high correlation between the latent variable $z$ and the conditional label variable $y | G(z)$ given an image $G(z)$ generated from $z$. Therefore, we can learn a linear mapping between the 2 spaces. Upon further investigation, we found out that the high correlation of variables in label space induces highly correlated coefficients of the mapping, results in the attempt in changing one attribute will consequently change other related attributes. In order to alleviate this phenomenon, we apply the orthogonality regularization \cite{bansal2018gain} into the loss function of the mapping to penalize similar coefficients.
Our main contributions are summarized as followed:
\begin{itemize}
\item[(i)] We propose a supervised method to explicitly map the latent space with a meaningfully pre-defined semantic space. In contract to other approaches; which often result in the \textit{difficulties in training}, \textit{limited number of properties} of the latent space that one can interpret and manipulate, and the \textit{accuracy} of manipulation; our method attempt to resolve all those demerits with the help of supervision. Instead of worrying about training a new GAN model (unlike InfoGAN), our model leverages of state-of-the-art GAN models to be responsible for generating most realistic images, then use simple model to perform the mapping and manipulation. This ease in implementing will allow many commercial applications to achieve competing results without too much cost of training.

\item[(ii)] We provide a method reducing the effect of high correlation in the label space, resulting in more robust image manipulation. For instance, attractiveness is usually correlated with youth, heaviness of make-up, smiling,... Therefore, changing one attributes often results in the change of others. Reducing that effect will produce more robust manipulation. 

\end{itemize}

\section{RELATED WORKS}

\textbf{Generative Adversarial Networks (GAN)} \cite{goodfellow2014generative} decouple the quest of maximizing likelihood of images which looks real (given a distribution assumption) and creating unseen high quality images, to only focus on the later objective. The network relies on the competition of the generator, which generates images from a random noise vector that should look realistic enough to fool the discriminator, whose main job is to distinguish real and fake images. Due to its objective, GAN is able to sample the most realistic images compared to other approaches. 

% GAN’s training difficulties and PGGAN.

Training GAN is an tremendously challenging task, suffering from 4 main difficulties: non-convergence and instability~\cite{salimans2016improved}, mode collapse~\cite{berthelot2017began, metz2016unrolled},  unstable generator gradient~\cite{arjovsky2017principled}, and highly sensitive to hyper-parameter tuning. PGGAN~\cite{karras2017progressive} overcomes these obstacles with a break-and-conquer method, progressively training the generator and discriminator from low-resolution to high, resulting in faster and more stable training. In order to fight mode collapse, they use a simpler version of mini-batch discrimination and some other training techniques to discourage the so-called ``cat-and-mouse game'' of the generator and discriminator. With those techniques implemented, PGGAN produces the most realistic images compared to other algorithms.

% 2.2 GAN difficulties and PGGAN

% 2.3 GAN latent space interpretation.

\textbf{GAN latent space interpretation}. An important point should be cleared out is that the latent space is structured by the generator, it itself has no meaning.
Most researches about GAN’s latent space interpretation are based on unsupervised manner. For instance, paper \cite{radford2015unsupervised} investigate the transition between two images by sampling a series latent vectors, which lies in a linear path connect two original random points,  from a prior distribution. For example, an image of a man gradually transits to images of a woman through the above interpolation. 

The method in \cite{radford2015unsupervised} also employs the supervision of human classified images, attempts to show that latent space encodes meaningful semantics by using vector arithmetic. In particular, the mean of random vectors generating images of smiling women minus that of neutral women plus one of neutral men results in a vector that generates an image of a smiling man. 
 
One novel idea in \cite{chen2016infogan} is that instead of interpreting a fixed mapping, authors tried to create a controllable, interpretable mapping from the latent space to image space. Intuitively, they used another easily, semantically understandable hidden variable $c$ in a substantially lower dimension to encode the latent variable~$z$. By control properties of $c$, they gain access to learn controllable, disentangled representation $z$ of the image distribution.

\textbf{Main drawbacks of other approaches to interpret GAN's latent space}.
Overall, despite the potential of in learning controllable, interpretable latent space, InfoGAN and other referred methods in this paper suffer from the limited number semantics that they can interpret. Our approach offers a richer set of attributes, because of its supervised manner, the number of semantics we can manipulate depends on the number of attributes we can label. Note that, the desired human-label attributes should have as low correlation as possible.

Besides, difficulties in implementation and training (InfoGAN), lack of preciseness and robustness in manipulation are also major drawbacks in many other methods. Our method will enable us to resolve in a simpler manner.

\section{PROPOSED METHODS}
\label{method}
\subsection{Classification model}
We first utilize ResNet-34 \cite{he2015deep} model pre-trained on ImageNet \cite{imagenet_cvpr09} as the backbone of our classification model. We then append the concatenation of adaptive average pooling and adaptive max pooling layers into the backbone, with the addition of 2 more dense blocks at last (each contains a dense layer, followed by Batch normalization, ReLU activation, and dropout). The model outputs a tensor in $\mathbb{R}^{40}$ with each element from [0, 1]  for every input image, indicating 40 predictions for 40 annotated attributes, where values near 0 indicate that the image almost does not have that attributes, and vice versa. We fine-tune the model with resized input images of size $128\times128\times3$ then use those weights to fine-tune the same model again with input size of $256\times256\times3$. This method enhances the accuracy of the model and will be described in more details in the training section.

With the aid of residual connection, we can increase the capacity of our model without the performance on training set getting hurt. We manage to prevent the over-fitting issue, induced by using ResNet-50 instead of ResNet-34, with heavy data augmentation and the same method on the input image size - which we used when training model with ResNet-34 as backbone. The promising result is further discussed in our Experiments section.

\subsection{Linear mapping from latent space to semantic space}
Interpreting the latent space in unsupervised learning is challenging, so we take advantage of the well - interpretable label space, to aid our understanding of the hidden space. We hypothesis that the hidden space can be compressed thus represented by another hidden space in lower dimension.
\begin{figure*}[t]
  \centering
  \includegraphics[height=25mm]{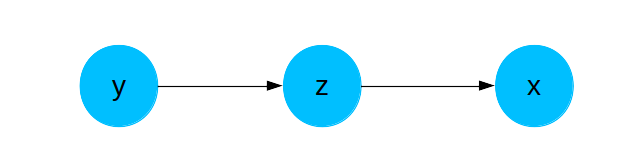}
  \caption{Probabilistic Graphical Model. The graph indicate a dependent relationship between variables}
  \label{fig1}
\end{figure*}

\begin{figure*}[t]
  \centering
  \includegraphics[width=1\linewidth]{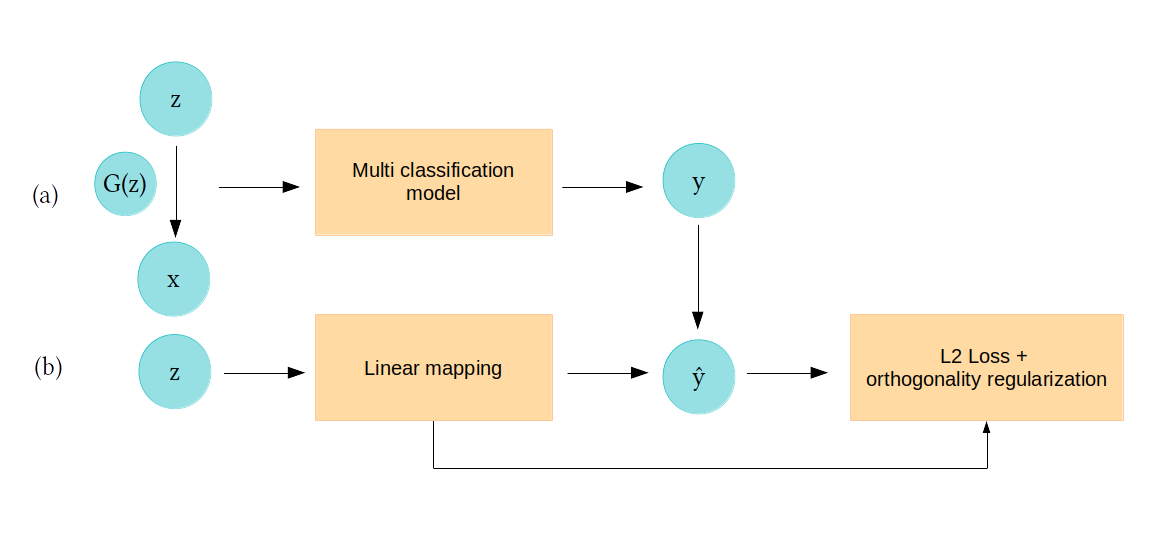}
  \caption{Proposal Model Architecture}
  \label{fig2}
\end{figure*}

 Since both latent and label variables can be perceived as factors explaining the data’s variation, we expect the label variable $y$ to have high correlation with the latent variable $z$. Hence, our goal is to approximate the relationship between the 2 variables by a linear mapping. 

\begin{alignat}{3}
y = z W^\top + b, 
\end{alignat}
% 
% __matrix form__
where $W$, $b$ is the coefficient matrix and intercept term, respectively.

Our main quests are how to achieve a change in a latent vector gives a corresponding change in the generated image and how to change one attribute without changing others. In other words, for example, how can we make the girl in a generated image look younger without changing her hair or even her gender.

$y_i$ is the attribute that we want to change by an amount of alpha, $\alpha$. Gen-image is the result of passing random latent vector $z$ through generator of GAN model. For instance, if $y_i$ indicates attractiveness feature of the image and alpha is positive, we hope that the new latent vector:
   \( z’ = z + \alpha w_i\)
  will result in a generated image of the same person but looks younger.

\begin{alignat}{3}
    z' = z + \alpha w_i \label{eq2}\\
    y' = z' W^\top + b
    \label{eq3}
\end{alignat}
\begin{alignat*}{3}
    \Leftrightarrow y + \Delta y &= (z + \alpha w_i) W^\top + b \\
    \Leftrightarrow y + \Delta y &= (z W^\top + b) + \alpha w_i W^\top \\
    \Leftrightarrow 
    \begin{bmatrix}
      \Delta y_1 \\
      \Delta y_2 \\
      \vdots \\
      \Delta y_n
    \end{bmatrix}
    &= \alpha w_i 
    \begin{bmatrix}
      \Delta w_1^\top \\
      \Delta w_2^\top \\
      \vdots \\
      \Delta w_n^\top
    \end{bmatrix}
    = \begin{bmatrix}
      \Delta \alpha w_1^\top w_i \\
      \Delta \alpha w_2^\top w_i\\
      \vdots \\
      \Delta \alpha w_n^\top w_i
    \end{bmatrix}\\
    \Rightarrow \Delta y > 0
\end{alignat*}

However, the formula above pointed out that if $w_i$ is `$similar$’ to other coefficients vector (measure by their cosine distance), change $z$ to $z’$ does not only manipulate $y_i$ attributes but also causing changes in other attributes. Consequently, our desire is to make the person in the original generated image look prettier can end up with a totally different person who is actually more attractive.  

In order to disentangle the linear mapping, we place an orthogonal penalty on the coefficient matrix. 
Therefore, the loss is computed as follows:

\begin{equation}
    J(w) = \text{MSE}(f_w(z), y) + \lambda \Vert w^\top w - I \Vert, 
\end{equation}
\text{where} $f_w \in \mathbb{R}^{512 \times 40}$ \text{linearly transforms $z$ to $y$}\\ \text{$\lambda$ is parameter coefficient, $I$ is Identity matrix}

The effect of this added regularization will be discussed in details in the Experiment section~\ref{experiments}. 

\section{Experiments \& Results}
\label{experiments}
\subsection{Dataset}

CelebFaces Attributes Dataset (CelebA) \cite{liu2015faceattributes}  is a large-scale face attributes dataset contains 10,177 celebrities images, each of which has 20 images. CelebA has large diversities, large quantities, rich annotations, including 202,599 number of face images, 5 landmarks locations, and 40 face attributes annotated by professional labeling company.

\subsection{Classification model results}
Beside the architecture design described in the above section, we apply several training techniques in order to boost up the performance of the multi-label classification model. In this part, we refer to FastAI \cite{howard2018fastai} as the main source of inspiration. 

Firstly, we set our learning rate schedule to a cyclical learning rate, also called the one-cycle policy \cite{smith2018disciplined}. If the learning rate is too small, mini-batch gradient cannot help parameters escape from an undesirable, narrow local minimum at early stage of training. Therefore, a practice of warming up the gradient (increase the gradient at first then decrease it later in the training course) can boost up learning at the beginning. 

Additionally, learning rate discrimination is applied to fine-tune the model. This makes sense as lower convolutional layers detect low-level features of an image such as edges, curves… and those features for natural images are more or less the same despite of the fact that whether its the image of a car or a human face. However, higher convolutional layers detects more conceptual texture such as the existence of a tail or car headlights or the color of a human eyes. Consequently, their weights need to be updated more when training in a new dataset. 

We use heavy augmentation as one of the main techniques to prevent over-fitting including horizontal flipping, wrapping, rotating but not cropping or brightening because these two transformations can partially delete or mutate important information of the image relating to attributes prediction.

Lastly, we apply a simple methodology of progressively multiple input size training as described in section \ref{method} above. We experiment that training with this technique gives a slightly better result with a little help of reducing over fitting than starting with images size of $256\times256\times3$ in the first place. The technical reason that we are able to implement this technique is based on our usage of adaptive pooling layers, which enable us to obtain fixed output size despite various input sizes.

\begin{table}[t]
\caption{Multi-label classification model's results}
\centering
 \begin{tabular}{c c c c c c} 
 \toprule
  \textbf{epoch} & \textbf{train\_loss} & \textbf{val\_loss} & \textbf{acc\_thres} & \textbf{fbeta} & \textbf{time (mm:ss)}  \\ 
  \midrule
  14 &0.207458 &0.190790 &0.916207 &0.843260 &04:19\\
  15 &0.193289 &0.182742 &0.919622 &0.848778 &04:20\\
  16 &0.188012 &0.179203 &0.921217 &0.851056 &04:21\\
 \bottomrule
 \end{tabular}
\label{mean_iou3}
\end{table}

After 16 epochs of early stopped fine-tuning, our classification model ends up with 92.1\% of accuracy and 0.85 F1 score. 

\subsection{Linear mapping and attributes manipulation}

\subsubsection{Overall setting of experiments}

At first, we sample 3000 random noise vectors from multivariate standard Gaussian distribution. We then use our classification model to obtain 3000 label vectors, each contains 40 attributes, from 3000 generated images using these above random noises (Note that, each elements of each label vector will be in [0, 1] instead of {0, 1}).

Its worth $p$ out a non-trivial implementation detail is that we generate $512\times512\times3$ images for each random noise vector in $\mathbb{R}^{512}$, then feed to the classification model a resized $256\times256\times3$ images, instead of generating $256\times256\times3$ images from random vectors in $\mathbb{R}^{256}$. One reason is that the random noise in $\mathbb{R}^{512}$ allows mapping to a higher qualification image. 

Secondly, we apply a linear transformation to approximate the mapping between the latent space $z$ and label space $y$. In order to increase cosine distance between coefficients vectors, we add the orthogonality regularization to the loss with weight $\lambda$ equal 2. The result is robust to this choice of $\lambda$ considering many trials.

\begin{figure*}[t]
    \centering
    \includegraphics[width=\linewidth]{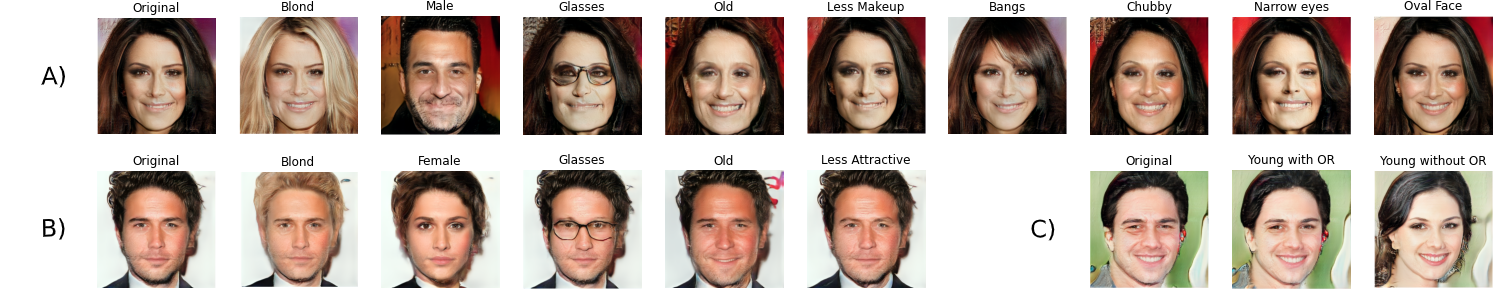}
    \caption{Image Manipulation with latent vector }
    \label{fig3}
\end{figure*}

\subsubsection{Analyze results}

%Figure … shows the manipulations of generated images using formula (\ref{eq2}) and (\ref{eq3}) with and without penalty for high-cosine-similarity coefficients. (describe the figure here). 
%Figure … shows the change in predicted attributes of images after being manipulated. (describe more about the figure)
Fig \ref{fig3} shows the manipulations of generated images using formula (\ref{eq2}) and (\ref{eq3}) with and without penalty for high-cosine-similarity coefficients.

\begin{table*}[t]
\caption{Changes in how manipulating young attributes affects other attributes, with and without Orthogonality Regularization}
\centering
\begin{tabularx}{\textwidth}{@{}l*{6}{C}c@{}}
%  \begin{tabular}{| c | c | c | c | c | c | } 
 \toprule
  \textbf{} & \textbf{Original} & \textbf{tfm. w/o reg.} & \textbf{abs\_diff\_no\_reg} & \textbf{tfm\_attr\_reg} & \textbf{abs\_diff\_reg}  \\ [0.5ex] 
 \midrule
  \textbf{Male} &0.999618 &0.000217 &0.999401 &0.997077 &0.002541\\
  \textbf{Makeup} &0.001739 &0.999788 &0.998050 &0.009618 &0.007879\\
  \textbf{Lipstick} &0.001942 &0.999757 &0.997815 &0.014711 &0.012769\\
  \textbf{Earrings} &0.030412 &0.814261 &0.783849 &0.088462 &0.058051\\
  \textbf{BagsUnderEyes} &0.782150 &0.027579 &0.754571 &0.671357 &0.110794\\
  \textbf{WavyHair} &0.118599 &0.870955 &0.752356 &0.190311 &0.071713\\
  \textbf{5oClockShadow} &0.688798 &0.000454 &0.688344 &0.672848 &0.015950\\
  \textbf{BigNose} &0.750373 &0.073389 &0.676984 &0.643183 &0.107190\\
  \textbf{OvalFace} &0.337116 &0.944806 &0.607690 &0.531252 &0.194136\\
  \textbf{DoubleChin} &0.542601 &0.001489 &0.541111 &0.3252063 &0.290538\\
  \textbf{BushyEyebrows} &0.960830 &0.421305 &0.539525 &0.972407 &0.011577\\
    \textbf{Young} &0.612252 &0.991196 &0.378944 &0.896836 &0.284584\\
    \textbf{ArchedEyebrows} &0.89847 &0.463946 &0.374099 &0.065678 &0.024169\\
    \textbf{NarrowEyes} &0.391831 &0.067585 &0.324246 &0.296349 &0.095482\\
  \bottomrule
 \end{tabularx}
\label{compare_table}
\end{table*}

With orthogonality regularization, increasing the Young attributes only makes the man look younger, unlike without the regularization, the manipulation causes change in many other properties including gender. Take a closer look at table Table \ref{compare_table}, where we pass original generated; manipulated with and without penalties images through the multi-label classification model and observe the change in their confidence score, we recognize that, increasing younger changes the man’s gender, makes him wear heavy makeup, lipstick and other related attributes. While comparing to the changes in those same attributes, the table indicates that with the help of regularization, making the man look younger relatively does not cause severe effect on other properties. We observe that same phenomenon in manipulating many other images.

\begin{table}[h!]
\caption{Compare cosine distance of Young and other attributes, with and without Orthognality Regularization}
\centering
 \begin{tabular}{ c  c  c } 
 \toprule
  \textbf{} & \textbf{Without Regularization} & \textbf{With Regularization}  \\ [0.5ex] 
 \midrule
  \textbf{Young} &1.000000  &1.000000\\
  \textbf{Attractive} &0.859559  &0.313507\\
  \textbf{HeavyMakeup} &0.795307  &-0.002151\\
   \textbf{Lipstick} &0.791769  &0.282624\\
   \textbf{NoBeard} &0.599438  &0.262185\\
   \textbf{BigLips} &0.542247  &-0.000376\\
   \textbf{ArchedEyebrows} &0.513396  &0.042719\\
   \textbf{OvalFace} &0.453567  &-0.017463\\
   \textbf{WavyHair} &0.430066  &0.036656\\
   \textbf{PointyNose} &0.413676  &0.022878\\
   \textbf{DoubleChin} &-0.786877  &0.019917\\
   \textbf{Male} &-0.769961  &0.039103\\
   \textbf{BagsUnderEyes} &-0.754783  &0.015170\\
   \textbf{GrayHair} &-0.749704  &0.014643\\
   \textbf{BigNose} &-0.734693  &0.034192\\
    \textbf{WearingTie} &-0.669467  &0.038437\\
    \textbf{Bald} &-0.585021  &0.042678\\
    \textbf{RecedingHairline} &-0.523914  &0.009939\\
    \textbf{Goatee} &-0.495711  &0.038485\\
 \bottomrule
 \end{tabular}
\label{cosine_distance_table}
\end{table}

For numerically reasoning why, Table \ref{cosine_distance_table} compares the cosine distances between Young and other semantics’ coefficients. Coefficients of Young is highly similar to those of attributes, a heavy makeup, increasing Young also increases those attributes. The opposite happens to Male, Chubby, or Bald, and more. Intuitively, being bald makes us look older. However, orthogonality regularization successively prevents these changes. It makes Young’s coefficient much less ‘similar’ to others’. The same observation shares among other attributes’ coefficients.

Fig \ref{fig3} also demonstrates the success of our approach (1st and 2nd row) through examples. Additionally notice that, women generally have more attributes that can be independently manipulated than men, typically heavy makeup, oval face, big lips… We suspect that the natural differences in men's and women's appearance and taste make the input data-set imbalance which has fewer men wearing makeup and possess oval face than women and more.

\subsection{Applications}
With the ease of implementation, accuracy, robustness, and flexibility in image manipulation, we believe that our model can boost up and enrich the performance of several Face Editing, Age Editing and Photo Makeover applications in the market.

\subsection{System configuration}
Our experiments are conducted on a computer with Intel Core i5-7500 CPU @3.4GHz, 32GB of RAM, GPU GeForce GTX 1080 Ti, and 1TB SSD hard disk. The models are implemented with the PyTorch \cite{pytorch} framework.

\section{Conclusion and future work}

%We proposed an new supervised method to map the latent space with a pre-fixed meaning, interpretable semantic space and a method reducing the effect of high correlation in the label space, result in more robust image manipulation.
Our model proposal utilizes human prior knowledge to learn the mapping of a rich set of meaningful, interpretable semantics to GAN’s latent space. Besides its main advantages as the ease in learning the mapping using our model and the number of attributes it is able to manipulate, our model suffers from a major drawback. That is the expense of labeling due to the supervised manner. In the future, we would like to combine the unsupervised-based Data labeling tool developed by FastAI to tremendously aid the process of data annotation. 

% \begin{alignat}{3}
% y &= x \begin{bmatrix}
%   \vec{w_{1}^\top} \\
%   \vec{w_{2}^\top} \\
%   \vdots \\
%   \vec{w_{i}^\top}
% \end{bmatrix}
% + b
% \end{alignat}

% \begin{alignat}{3}
% z' = z+ w_i \alpha
% \end{alignat}

\section*{Acknowledgment}
This work is partially supported by \textbf{\textit{Sun-Asterisk Inc}}. We would like to thank our colleagues at \textbf{\textit{Sun-Asterisk Inc}} for their advice and expertise. Without their support, this experiment would not have been accomplished.

\bibliographystyle{IEEEtran}
\bibliography{references}

\end{document}